# Development of a dynamic type 2 diabetes risk prediction tool: a UK Biobank study

**Authors:** Nikola Dolezalova[1]; Massimo Cairo[1]; Alex Despotovic[1,2]; Adam T.C. Booth[1]; Angus B. Reed[1], Davide Morelli[1,3]; David Plans[1,4]

1. Huma Therapeutics Limited, London, United Kingdom
2. Faculty of Medicine, University of Belgrade, Belgrade, Serbia
3. Department of Engineering Science, Institute of Biomedical Engineering, University of Oxford, Oxford, United Kingdom
4. University of Exeter, SITE, Exeter, United Kingdom

**\* Corresponding Author:** David Plans (david.plans@huma.com)

# Abstract

Diabetes affects over 400 million people and is among the leading causes of morbidity worldwide. Identification of high-risk individuals can support early diagnosis and prevention of disease development through lifestyle changes. However, the majority of existing risk scores require information about blood-based factors which are not obtainable outside of the clinic. Here, we aimed to develop an accessible solution which could be deployed digitally and at scale. We developed a predictive 10-year type 2 diabetes risk score using 301 features derived from 472,830 participants in the UK Biobank dataset, while excluding any features which are not easily obtainable by a smartphone. Using a data-driven feature selection process, 19 features were included in the final reduced model. A Cox proportional hazards model slightly overperformed a DeepSurv model trained using the same features, achieving a concordance index of 0.818 (95% CI: 0.812–0.823), compared to 0.811 (95% CI: 0.806–0.815). The final model showed good calibration. This tool can be used for clinical screening of individuals at risk of developing type 2 diabetes and to foster patient empowerment by broadening their knowledge of the factors affecting their personal risk.

# Introduction

Diabetes is one of the fastest growing causes of global morbidity, rising from the 20th most impactful disease in 1990 to the 8th in 2019 (1). Diabetes risk scores are an important tool in the identification of patients at risk of developing type 2 diabetes (T2D), aiding clinical decision making and patient support. A broad range of diabetes risk scores have been developed (2), some of which are embedded in clinical practice guidelines (3,4). The widespread use of current T2D risk scores is limited by the requirement of clinical or laboratory tests, not routinely available to patients (2). These necessitate a visit to the hospital or GP, thereby reducing the accessibility of risk profiling.

One solution may be screening, in a remote setting, through smartphone apps. This method, along with easily assessed variables could facilitate patient screening at scale. Provided that the tool used for screening has strong evidence to support its use, earlier identification of high-risk patients and possible mitigation through adequate lifestyle and behavioural modification—known to be the primary drivers of T2D prevention—will be possible (5).

The primary aim of this study was to develop a model to predict the 10-year risk of T2D using the UK Biobank (UKB) cohort, considering only factors that could be captured via smartphone in a remote setting. The secondary aim was the identification of novel features which could enrich traditional risk-profiling.

# Methods

Data for this study was derived from the UKB dataset, whose participants were recruited between 2006 and 2010 from the UK general population (6). Use of this data for our study was approved by the UKB under application number 55668.

The definition of outcome for this study was an incident T2D. The date of diagnosis was obtained from the UKB 'First Occurrences' fields for ICD10 codes E11–E14 (contains information from primary and secondary care, death register, and self-report). Participants with type 1 diabetes (diagnosed with ICD10 code E10) or pre-existing at the date of enrolment in UKB T2D (E11–E14) were excluded from the study. Censoring was applied at the time of first T2D diagnosis, date of death, date participant was lost to follow-up, or the end date of the study (30th September 2020).

A set of UKB variables, selected based on clinical insight and accessible via smartphone, were used as inputs and included: demographic characteristics, anthropometric measures, lifestyle measures, medical history, and family history. Participants with any missing records were excluded from the study. Categorical and ordinal features were one-hot encoded. Test data (25%) remained unseen during training and was used for internal validation of the models.

An initial baseline model was built using a Cox proportional hazards (CPH) model using an implementation from the python lifelines library (7). Baseline features were then subjected to feature selection using stepwise backwards elimination, whereby only features whose exclusion led to degradation of the concordance index (c-index) more than a standard deviation obtained during 2-fold cross-validation were kept in the final reduced set. Shortlisted features were reviewed by clinical consensus (*Table 1*).

In addition to the CPH model, we trained a Cox proportional hazards deep neural network (DeepSurv), implemented in the 'pycox' package (8). Hyper-parameter searching was performed for the reduced set of features (search space and selected hyperparameters are described in *Supplementary Table 1*).

C-index was used as a measure of discrimination for both CPH and DeepSurv models on test dataset. Confidence intervals (CIs) were calculated using bootstrapping with 50 resampling rounds. Calibration of the models for 10-year predictions was evaluated using the Python lifelines library (7). In the demographic summary, groups were compared using Chi-squared test for categorical and ordinal variables and Kruskal-Wallis test for continuous variables.

# Results

The final datasets, after applying exclusion criteria, contained 472,830 participants, 4.03% of which developed T2D during the observation period. Details of participant flow through the study can be found in *Supplementary Figure 1*. Median survival time was 11.2 years (interquartile range 10.8–12.3 years). Median age of participants was 58, and 55.4% of the participants were female (summary of the distribution of variables in the final model can be found in *Supplementary Table 2*).

The initial 88 variables were expanded to 301 features after one-hot-encoding. The baseline CPH model, containing 301 variables, achieved a concordance index of 0.825 (95% CI: 0.821–0.832). After backward elimination and clinical refinement, feature space was reduced to 19. The remaining features are detailed in *Table 1*.

Feature reduction and refinement had minimal impact on model performance (c-index: 0.818, 95% CI: 0.812–0.823). Detailed results with feature coefficients can be found in *Figure 1* and *Supplementary Table 3*. The mean predicted 10-year risk was 3.59%, the mean observed 10-year risk was 3.29%. The Integrated Calibration Index (ICI) was 0.298%, suggesting good calibration of the model. When a DeepSurv model was trained with the same set of features using optimised model hyperparameters, it achieved a similar c-index of 0.811 (95% CI: 0.806–0.815).

# Discussion

Using the UKB dataset, we developed a simple, accessible risk model for long-term prediction of T2D. Our model showed good discrimination, with c-index of 0.82, and good calibration. By using only inputs obtainable through a smartphone, this model could allow personalised risk assessment without the presence of a clinician.

If implemented in a smartphone application, our risk model could serve as a valuable tool for the early detection of individuals at high risk of developing T2D. It could also help to foster informed patient autonomy regarding their own health by highlighting protective and harmful factors, alongside ways to manage personal risk through lifestyle alterations. To increase user engagement with such a solution, efforts could be made to gamify the implementation of this model through features such as goal setting within a smartphone application. Emerging technology for body measurement acquisition using a smartphone camera (9), or passive tracking of activity (10) or sleep data from wearables, offer an opportunity to both streamline and diversify input collection, further improving user engagement and dynamic risk modelling.

Diabetes risk scores that use results of either blood glucose or glycated haemoglobin tests (3,11) or history of high blood glucose (12) generally achieve better c-indices than those which do not use these measures. Our model, however, overperforms several others which do not require previous blood tests, such as the Leicester Risk Assessment Score with AUC-ROC of 0.72, the Sri Lanka Diabetes risk score (SLDRISK) with 0.78 or the web-based ADA score with 0.77 in external validation (4,13,14). In comparison to the non-blood-based QDScore, developed using the QRESEARCH database and used in clinical practice, similar discrimination was achieved (0.83 vs 0.82) despite QDScore comprising a lower-risk population. Our model further includes variables that could be easily incorporated into existing clinical assessments as more dynamic indicators of risk.

Many features in the final model are traditional T2D risk factors, such as age, ethnicity, BMI, smoking, and family history of diabetes (15). Interestingly, exclusion of sex from the final model features did not lead to degradation of the c-index and was, therefore, excluded from the reduced model. In line with the findings of some other studies, alcohol consumption was shown as protective in our model (16). Among the novel factors are self-reported overall health and some self-reported symptoms, such as stomach or abdominal pain, shortness of breath, and daytime sleeping. While these symptoms are all more common in individuals diagnosed with diabetes (17–19), their use in risk evaluation tools has not been reported. They could be signs of undiagnosed diabetes or other comorbidities strongly associated with T2D, such as cardiovascular diseases or subclinical metabolic conditions.

The results of this study should be viewed with respect to the limitations conferred by the dataset. UKB participants are on average healthier and wealthier than the general UK population, and the cohort is unrepresentative regarding ethnic background (20). By foregoing a data-driven approach, additional potentially useful variables present in the dataset may have been missed during the initial clinical selection process. Incidence of T2D may have been missed, as detailed screening of primary care records was not available. External validation of this model is required prior to its implementation within demographically dissimilar populations.

To conclude, we developed a simple 10-year T2D risk model using dynamic and smartphone-accessible variables from the UKB dataset with very good discrimination and calibration metrics.

**Acknowledgements:** This research was funded by Huma Therapeutics Ltd. The funders had no role in study design, data collection and analysis, decision to publish, or preparation of the manuscript. We would like to thank Michele Colombo for his contribution.

**Conflict of Interest:** N.D., M.C., A.D., A.T.C.B, A.B.R. D.M., and D.P. are employees of Huma Therapeutics Ltd.

**Figure legends**

*Figure 1: Plot of Cox Proportional Hazards model coefficients.* *Values show log(HR) ± 95% CI. HR = hazard ratio, CI = confidence interval.*

# Supplementary Materials

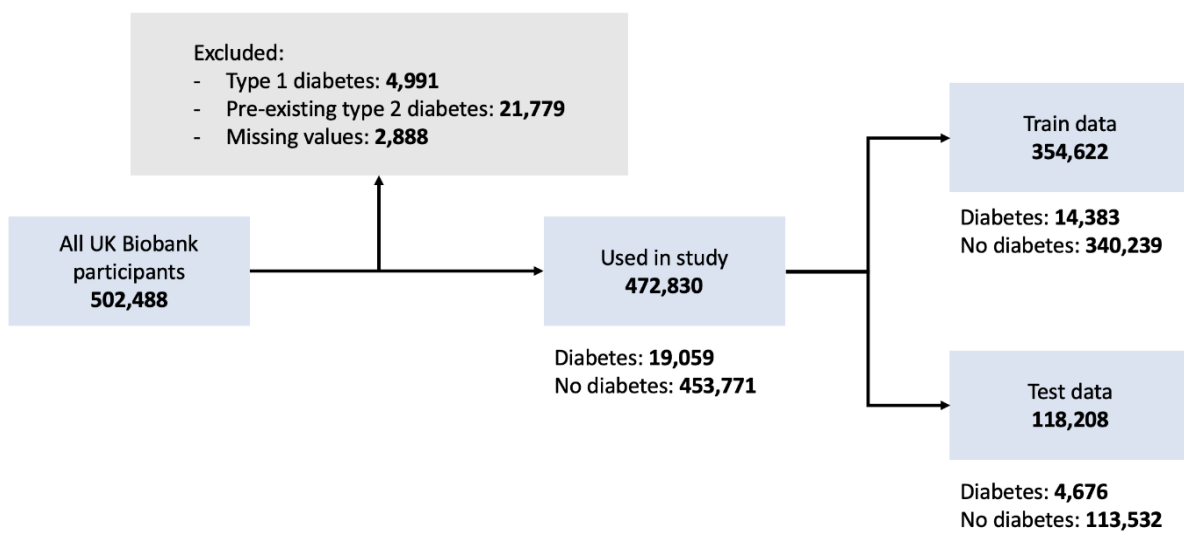

*Supplementary Figure 1: Flow diagram of participants through the study.* *After applying exclusion criteria, the data was split into train (75%) and test datasets (25%), stratified on outcome (incident type 2 diabetes). Breakdown of the diabetes incidence in the datasets is shown underneath the boxes.*

*Supplementary Table 1: DeepSurv hyperparameter search space and optimal parameters.* Tree-Structured Parzen Estimator algorithm[A] from the Optuna library[B] was used to find the optimal set of parameters within the search space.

| Hyper-parameter | Search space | Optimal for reduced model |
|---|---|---|
| Activation | LeakyReLU[C], ReLU[D], and SELU[E] | SELU |
| Hidden layers topology | 8, 32, 256, 32x32, 64x64, 128x128, 64x16, 256x32, 32x32x32, 64x64x64 | 64x64x64 |
| Drop-Out*[F] | [0, 0.9] | 0.04809 |
| Weight-Decay*[G] | [0, 20] | 0.00101 |
| Batch Normalization[H] | Yes/No | Yes |
| Optimizer | Stochastic Gradient Descent, Adam[J] | Adam |
| Momentum*[K] | [0,1] | Not used for Adam |
| Learning Rate | Log distribution on [1e$^{-5}$, 1] | 0.00169 |

*Uniform distributions*

*Supplementary Table 2: Cohort demographics of the final dataset.* Last column shows p-value after comparing the groups with different incident T2D outcomes. Binary variables were compared using the Chi-squared test, continuous variables using the Kruskal-Wallis test. T2D = type 2 diabetes, Q1 = 1st quartile, Q3 = 3rd quartile

|  | Overall | Without incident T2D | Incident T2D | P-Value (adjusted) |
|---|---|---|---|---|
| n | 472830 | 453771 | 19059 | |
| Age, median [Q1,Q3] | 58.00 [50.00,63.00] | 57.00 [50.00,63.00] | 60.00 [54.00,65.00] | <0.001 |
| Waist/hip ratio, median [Q1,Q3] | 0.87 [0.80,0.93] | 0.87 [0.80,0.93] | 0.94 [0.88,0.99] | <0.001 |
| BMI, median [Q1,Q3] | 26.57 [24.03,29.64] | 26.44 [23.95,29.42] | 30.59 [27.56,34.42] | <0.001 |
| College/university degree, n (%) | 153919 (32.55) | 149992 (33.05) | 3927 (20.60) | <0.001 |
| Ethnicity - Asian, n (%) | 9447 (2.00) | 8502 (1.87) | 945 (4.96) | <0.001 |
| Ethnicity - Black, n (%) | 7031 (1.49) | 6410 (1.41) | 621 (3.26) | <0.001 |
| Ever diagnosed Heart attack / Angina / Stroke / High blood pressure, n (%) | 131458 (27.80) | 121116 (26.69) | 10342 (54.26) | <0.001 |
| Medications for cholesterol, n (%) | 66807 (14.13) | 60606 (13.36) | 6201 (32.54) | <0.001 |
| Other prescription medications (excl. diabetes/cholesterol/blood pressure), n (%) | 213365 (45.13) | 200968 (44.29) | 12397 (65.05) | <0.001 |
| Stomach or abdominal pain in last month, n (%) | 40837 (8.64) | 38468 (8.48) | 2369 (12.43) | <0.001 |
| Narcolepsy, n (%) | 112170 (23.72) | 105384 (23.22) | 6786 (35.61) | <0.001 |
| Shortness of breath walking on level ground, n (%) | 16565 (3.50) | 14965 (3.30) | 1600 (8.39) | <0.001 |
| Diabetes in father, n (%) | 39837 (8.43) | 37505 (8.27) | 2332 (12.24) | <0.001 |
| Diabetes in mother, n (%) | 40972 (8.67) | 37951 (8.36) | 3021 (15.85) | <0.001 |
| Diabetes in siblings, n (%) | 30469 (6.44) | 27829 (6.13) | 2640 (13.85) | <0.001 |
| Drinks alcohol once a month or more, n (%) | 383373 (81.08) | 370239 (81.59) | 13134 (68.91) | <0.001 |
| Previous smoker, n (%) | 49592 (10.49) | 46759 (10.30) | 2833 (14.86) | <0.001 |
| Smoking pack-years, median [Q1,Q3] | 0.00 [0.00,6.50] | 0.00 [0.00,6.00] | 0.00 [0.00,20.50] | <0.001 |
| Good or excellent health (self-reported), n (%) | 358734 (75.87) | 348776 (76.86) | 9958 (52.25) | <0.001 |

*Supplementary Table 3: Results of the reduced Cox proportional hazards model.* The table displays coefficients = log(HR), their 95% confidence intervals, and -log2 (p-value), where p < 0.05 an null hypothesis states that the coefficient is equal to 0.

| Covariate | log(HR) | CI log(HR) lower 95% | CI log(HR) upper 95% | -log2 (p-value) |
|---|---|---|---|---|
| Ethnicity - Asian | 0.844 | 0.764 | 0.925 | 308.259 |
| Ethnicity - Black | 0.532 | 0.436 | 0.628 | 88.794 |
| Diabetes in mother | 0.489 | 0.443 | 0.535 | 313.265 |
| Waist/hip ratio | 0.440 | 0.423 | 0.458 | > 500 |
| Diabetes in siblings | 0.422 | 0.372 | 0.471 | 207.302 |
| BMI | 0.399 | 0.386 | 0.413 | > 500 |
| Diabetes in father | 0.385 | 0.334 | 0.436 | 161.204 |
| Ever diagnosed Heart attack / Angina / Stroke / High blood pressure | 0.368 | 0.330 | 0.405 | 265.184 |
| Medications for cholesterol | 0.285 | 0.244 | 0.325 | 142.329 |
| Currently smoking | 0.278 | 0.229 | 0.328 | 90.997 |
| Other prescription medications (excl. diabetes/cholesterol/blood pressure) | 0.250 | 0.212 | 0.287 | 128.297 |
| Age | 0.242 | 0.222 | 0.262 | 409.902 |
| Stomach or abdominal pain in last month | 0.177 | 0.125 | 0.228 | 35.703 |
| Narcolepsy | 0.176 | 0.141 | 0.211 | 72.542 |
| Smoking pack-years | 0.086 | 0.074 | 0.098 | 145.292 |
| Shortness of breath walking on level ground | 0.031 | -0.031 | 0.093 | 1.634 |
| College/university degree | -0.217 | -0.258 | -0.175 | 78.980 |
| Good or excellent health (self-reported) | -0.323 | -0.360 | -0.286 | 212.648 |
| Drinks alcohol once a month or more | -0.375 | -0.413 | -0.337 | 276.673 |